%% file: acl_latex.tex
\documentclass[11pt]{article}

\usepackage[preprint]{acl}

\usepackage{times}
\usepackage{latexsym}

\usepackage[T1]{fontenc}

\usepackage[utf8]{inputenc}

\usepackage{microtype}

\usepackage{inconsolata}

\usepackage{graphicx}

\usepackage[utf8]{inputenc} 
\usepackage[T1]{fontenc}    
\usepackage{url}            
\usepackage{microtype}      
\usepackage{xcolor}         
\usepackage{xspace}
\usepackage{graphicx,float,subfig,subcaption}
\usepackage{amsmath,amssymb,amsfonts,dsfont,pifont,bm,mathrsfs,mathtools,nicefrac}
\usepackage{algorithm,algpseudocode,listings}
\usepackage{booktabs,multirow,adjustbox,diagbox,threeparttable}
\usepackage{makecell}
\definecolor{citeblue}{RGB}{48,111,186}
\usepackage[hypcap=false]{caption}
\usepackage{enumitem}
\usepackage{anyfontsize}
\usepackage{colortbl}
\usepackage{placeins}
\usepackage[most]{tcolorbox} 
\usepackage{minted}

\newtcolorbox{promptbox}{
    colframe=blue!70!black,   
    colback=white,            
    title=Prompt Example,     
    fonttitle=\bfseries,      
    boxed title style={colback=blue!70!black}, 
    attach boxed title to top center={yshift=-2mm}, 
    arc=4pt,                  
    boxrule=1pt,
    left=10pt,right=10pt,top=8pt,bottom=8pt
}

\setlist[itemize]{leftmargin=15pt}

\usepackage{listings}
\usepackage{xcolor}
\definecolor{codegreen}{rgb}{0,0.6,0}
\definecolor{codegray}{rgb}{0.5,0.5,0.5}
\definecolor{codepurple}{rgb}{0.5,0,0.5}
\definecolor{backcolour}{rgb}{0.95,0.95,0.92}
\definecolor{tableblue}{RGB}{218,242,250}

\lstdefinestyle{mystyle}{
    backgroundcolor=\color{backcolour},   
    commentstyle=\color{codegreen},
    keywordstyle=\color{magenta},
    numberstyle=\tiny\color{codegray},
    stringstyle=\color{codepurple},
    basicstyle=\ttfamily\footnotesize,
    breakatwhitespace=false,
    breaklines=true,                 
    captionpos=b,                    
    keepspaces=true,                 
    numbers=none, 
    numbersep=5pt,                  
    showspaces=false,                
    showstringspaces=false,
    showtabs=false,                  
    tabsize=2
}

\def\ie{\textit{i.e.}}

\newcommand{\method}{RouteMoA\xspace}
\definecolor{darkred}{rgb}{0.7, 0.0, 0.0}

\usepackage{adjustbox}
\title{RouteMoA: Dynamic Routing without Pre-Inference Boosts Efficient Mixture-of-Agents}

\author{
    Jize Wang$^{1} $\  Han Wu$^{1}$ \   Zhiyuan You$^{2}$  \  Yiming Song$^1$ \  Yijun Wang$^3$ \ Zifei Shan$^3$ \\
    \textbf{Yining Li}$^4$ \  
    \textbf{Songyang Zhang}$^4$ \  
    \textbf{Xinyi Le}$^{1*}$ \  
    \textbf{Cailian Chen}$^1$\thanks{Corresponding Authors.} \  
    \textbf{Xinping Guan}$^{1}$ \  
    \textbf{Dacheng Tao}$^{5}$ 
     \\
      $^1$ Shanghai Jiao Tong University \quad $^2$ CUHK \quad  
      $^3$ Tencent\\ $^4$Shanghai AI Laboratory \quad $^5$Nanyang Technological University
      \\  \texttt{\{jizewang2000,lexinyi,cailianchen\}@sjtu.edu.cn}
}

\begin{document}
\maketitle

\input{sections/0_abstract}
\input{sections/1_introduction_v1}

\input{sections/2_related_work}

\input{sections/3_method}

\input{sections/4_experiment}

\input{sections/5_conclusion}

\bibliography{custom}

\input{sections/7_appendix}

\end{document}

%% file: sections/0_abstract.tex
\begin{abstract}

Mixture-of-Agents (MoA) improves LLM performance through layered collaboration, but its dense topology raises costs and latency. Existing methods employ LLM judges to filter responses, yet still require all models to perform inference before judging, failing to cut costs effectively. They also lack model selection criteria and struggle with large model pools, where full inference is costly and can exceed context limits. 
To address this, we propose \textbf{RouteMoA}, an efficient mixture-of-agents framework with dynamic routing. It employs a lightweight \textit{scorer} to perform initial screening by predicting coarse-grained performance from the query, narrowing candidates to a high-potential subset without inference. A \textit{mixture of judges} then refines these scores through lightweight self- and cross-assessment based on existing model outputs, providing posterior correction without additional inference. Finally, a \textit{model ranking} mechanism selects models by balancing performance, cost, and latency.
RouteMoA outperforms MoA across varying tasks and model pool sizes, reducing cost by 89.8\% and latency by 63.6\% in the large-scale model pool.

\end{abstract}

%% file: sections/1_introduction_v1.tex
\section{Introduction}


Large Language Models (LLMs)~\cite{instructgpt,survey} demonstrate strong capabilities across diverse tasks. While general-purpose models (e.g., Llama-3.1~\cite{llama}, Qwen2.5~\cite{qwen}) show broad competence, specialized variants (e.g., Qwen2.5-Math~\cite{qwenmath}, Qwen2.5-Coder~\cite{qwencoder}) excel in specific domains. This diversity in expertise makes the effective integration of multiple LLMs a promising direction to achieve performance beyond individual models.

\begin{figure}[t]
    \centering
    \includegraphics[width=0.98\linewidth]{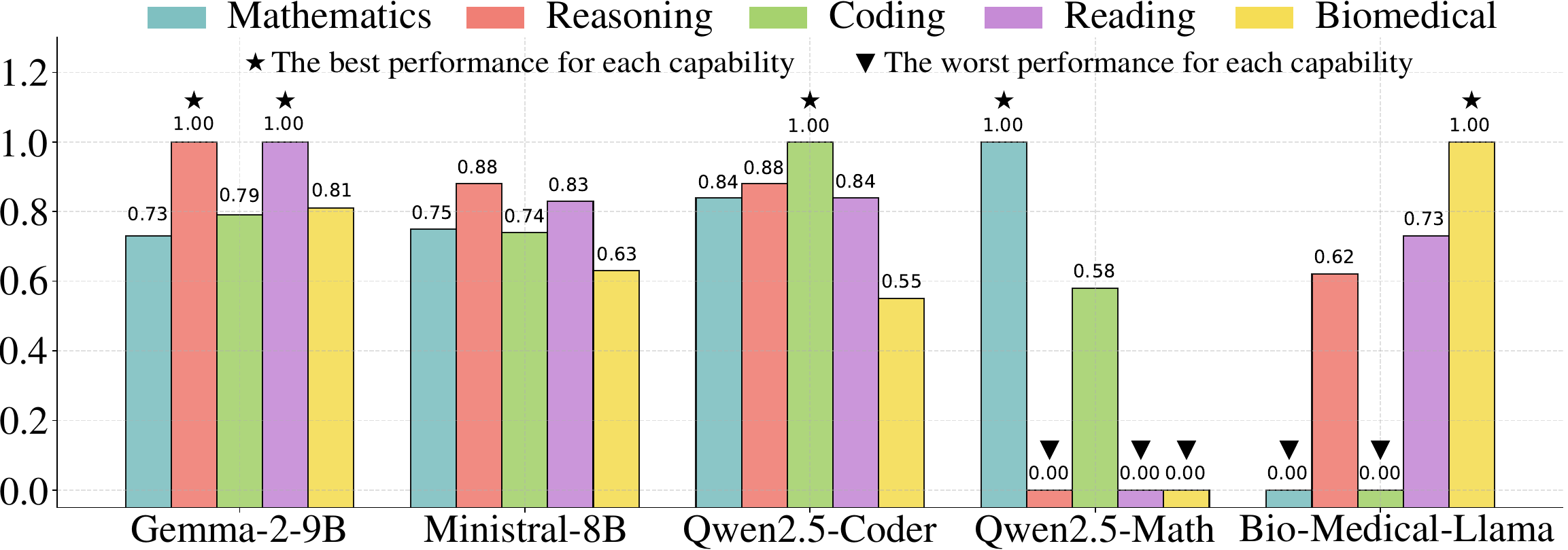}
    \caption{
     \textbf{Significant variations in model capabilities.} Values are normalized to [0,1]. Models exhibit clear specialization: Qwen2.5-Coder leads in coding but lags in biomedical tasks; Qwen2.5-Math excels in mathematics but struggles elsewhere; Bio-Medical-Llama dominates in biomedical knowledge but performs poorly in math and coding; Gemma stands out in reasoning and reading. These distinct profiles make it feasible to predict model performance only based on specific user queries.
    }
    \label{fig:teaser0}
    \vspace{-10pt}
\end{figure}

\begin{figure*}[t]
    \centering
    \includegraphics[width=0.9\linewidth]{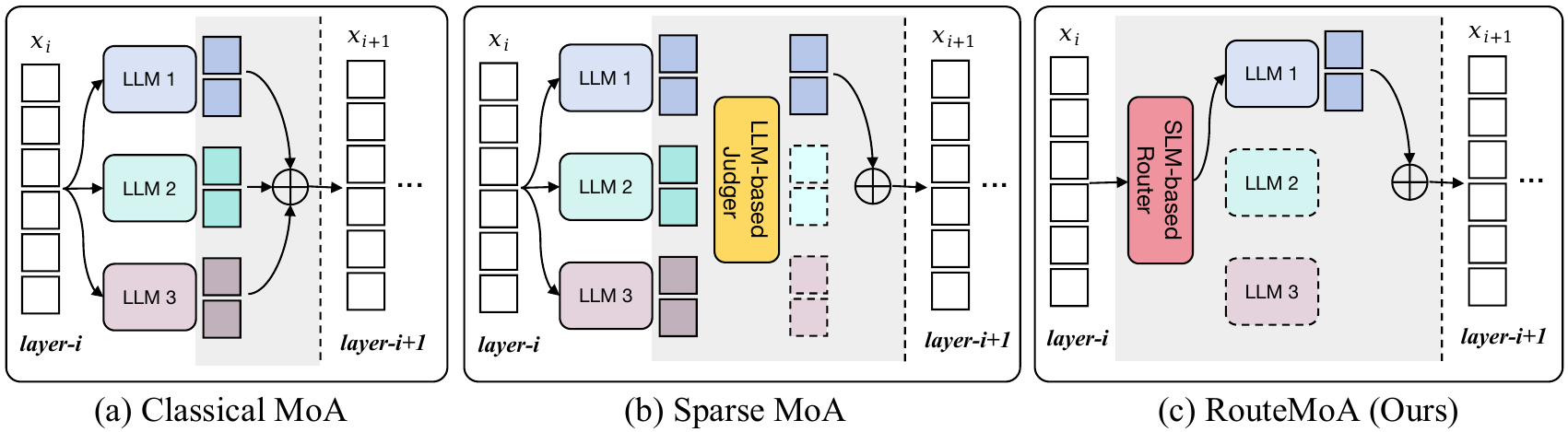}
    \caption{
    \textbf{Concept comparison} between our \method and previous MoA-based methods. (a) Classical MoA~\cite{moa} forwards all LLMs in each layer, and concatenates all outputs as the input of the next layer. (b) Sparse MoA~\cite{smoa} introduces an LLM-based judge to select some good responses as the input of the next layer. This reduces the number of input tokens, but still needs to forward all LLMs and another LLM-based judge. (c) \method uses a lightweight router to select parts of LLMs for inference, significantly reducing computational cost.
    }
    \label{fig:teaser}
    \vspace{-10pt}
\end{figure*}

Among various LLM-based multi-agent collaboration strategies, Mixture-of-Agents (MoA)~\cite{moa} is a typical and effective approach. As shown in Figure~\ref{fig:teaser}(a), this method enables multiple LLMs to refer to each other’s responses, engaging in iterative rounds (\ie, layers in Figure~\ref{fig:teaser}) of replies and summaries to achieve results superior to those of a single model. 

Despite the advantages, MoA-based methods are \textbf{\textit{highly resource-intensive}}. 
As shown in Figure~\ref{fig:teaser}(a), classical MoA~\cite{moa} requires forwarding multiple LLMs per layer and concatenating all outputs as the input to the next, leading to high cost and latency. 
Sparse MoA~\cite{smoa} (Fig.~\ref{fig:teaser}(b)) introduces a judge to filter responses, yet still invokes all LLMs plus an additional judge model, further increasing overhead. These approaches also lack principled model selection and do not scale to large pools (e.g., >10 models), as full inference becomes prohibitively costly and often exceeds context limits.

To address the efficiency challenge, we propose \method, a dynamically-routed mixture-of-agents framework. Our approach is motivated by the complementary capabilities of LLMs (Figure~\ref{fig:teaser0}): for example, Qwen2.5-Math excels in mathematics but underperforms in reasoning and biomedical tasks. Such specialization makes it feasible to predict model performance from the query, thus narrowing the initial pool to a few high-potential candidates and reducing cost.

Specifically, \method leverages a lightweight \textit{scorer} that performs initial screening. Using only prior knowledge from the query, it estimates model suitability without executing inference. It assigns coarse-grained scores to identify promising candidates, enabling activation of only a subset of models and significantly lowering inference overhead.

To correct potential scoring errors, we introduce a \textit{mixture of judges} to combine the scorer with self- and cross-assessment. These judges operate post-hoc, leveraging posterior knowledge from previously-generated responses without requiring additional inference. This design enhances assessment reliability at no extra cost, ensuring robust model selection throughout the routing process. Finally, a \textit{model ranking} mechanism selects models by balancing performance, cost, and latency.

In summary, our contributions are as follows:

\begin{enumerate}[label=\textbullet, leftmargin=12pt, topsep=0pt, itemsep=1pt, partopsep=1pt, parsep=1pt]

\item We propose RouteMoA, a dynamically-routed MoA framework that significantly cuts cost and latency while maintaining strong performance.

\item We design a lightweight scorer for initial model screening based on query-aware prior knowledge, narrowing the candidate pool to a few high-potential models without pre-inference.

\item We introduce a mixture of judges that refines model scores through self- and cross-assessment, leveraging posterior knowledge from model outputs to correct prediction errors without introducing additional inference overhead.

\item Extensive experiments on both small- and large-scale model pools, along with out-of-distribution tasks, show RouteMoA matches or surpasses strong baselines in accuracy while greatly boosting efficiency and scalability.

\end{enumerate}

%% file: sections/2_related_work.tex
\begin{figure*}[t]
    \centering
    \includegraphics[width=0.99\textwidth]{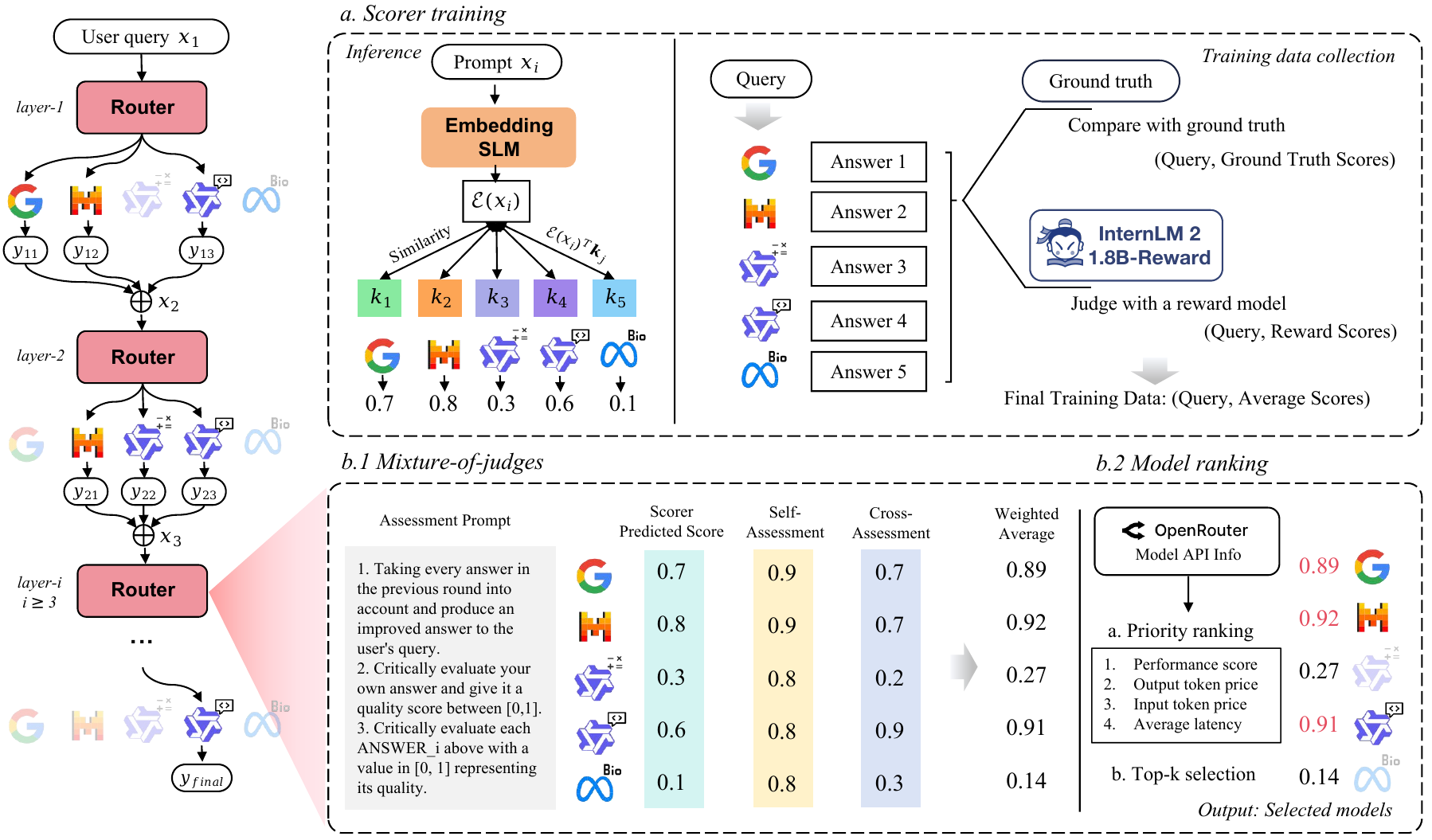}
    \vspace{-5pt}
    \caption{\textbf{RouteMoA architecture}. 
    The framework operates layer-wise (left). At each layer~$l$, the router selects a subset of suitable LLMs, whose outputs are aggregated and passed to the next layer. The router (right) consists of two stages: \textit{b.1 Mixture of Judges}, which includes a scorer (trained as in \textit{a. Scorer Training}), self-assessment, and cross-assessment. The scorer predicts candidate performance in layer-1 using prior knowledge from the query; subsequent layers refine scores via self- and cross-assessment using posterior knowledge from model outputs. \textit{b.2 Model Ranking} selects LLMs by balancing performance, cost, and latency.}
    \label{fig:construct}
    \vspace{-10pt}
\end{figure*}

\section{Related Work}

\textbf{General and task-specific LLMs.}
Large Language Models (LLMs) have shown strong performance in text understanding and generation~\cite{gpt4, internlm, llama, qwen}. They can be categorized into general-purpose models—such as Llama-3.1~\cite{llama3.1}, Qwen2.5~\cite{qwen}, Mistral~\cite{mistral}, and Gemma~\cite{gemini}—and domain-specific fine-tuned variants like Qwen2.5-Math~\cite{qwenmath}, Qwen2.5-Coder~\cite{qwencoder}, and Bio-Medical-Llama~\cite{biomed}. While specialized models excel in their domains, they often underperform elsewhere. For instance, Bio-Medical-Llama achieves 87.0 on the MMLU biomedical subset, outperforming Qwen2.5-Math by 43.5 points, yet scores only 11.7 on MATH. Since developing a universally capable model is costly, integrating multiple models’ strengths offers a more viable path.

\noindent\textbf{LLM-based multi-agent collaboration.}
Multi-agent frameworks provide effective ways to leverage diverse model capabilities. Majority voting~\cite{voting} selects the most frequent answer from multiple models as the final output. LLM cascading~\cite{llm_cascade} sequentially invokes models until a response meets a quality threshold. Multi-agent debate~\cite{multi_agent_debate} enhances accuracy through iterative discussion. Mixture-of-Agents (MoA)~\cite{moa} refines answers via multi-round parallel reasoning but incurs high computational cost. Sparse MoA~\cite{smoa} introduces a judge to filter responses, saving input tokens but still requiring inference from all models. In contrast, our approach adopts a lightweight router that selects suitable models dynamically for each layer without pre-inference, significantly cutting cost and latency.

\noindent\textbf{LLM routing.}
LLM routing~\cite{hybridllm,tensoropera_router} selects the best-performing model per query without invoking all candidates. Meta-models are trained to predict model performance based on input, improving cost efficiency. Benchmarks such as RouterBench~\cite{routerbench} and RouterEval~\cite{routereval} assess routing effectiveness. ZOOTER~\cite{zooter} distills reward signals into an SLM router via KL-divergence, while RouterDC~\cite{routerdc} uses dual contrastive loss for better accuracy. Eagle adopts a training-free approach using similarity-based retrieval. RouteLLM~\cite{routellm} focuses on binary routing between strong and weak models to minimize expensive calls. 
In contrast to routing methods that rely solely on query-based prior knowledge, our approach also leverages posterior knowledge from actual model outputs to update performance scores.
This design relaxes the requirement for precise performance prediction, and the subsequent multi-agent collaboration further enhances robustness and overall performance beyond what is achievable by routing to a single model.

%% file: sections/3_method.tex
\section{Methodology}\label{sec:methods}

In this section, we introduce RouteMoA, an efficient mixture-of-agents framework with dynamic routing. It dynamically selects a subset of top-performing LLMs each round without pre-inference, thus reducing cost and latency while maintaining performance. We first overview the whole routing process in Section~\ref{sec:overview}, then describe its key components: the scorer, mixture of judges, and model ranking (Sections~\ref{sec:scorer}–\ref{sec:ranking}).

\subsection{Overview}
\label{sec:overview}
The framework operates layer-wise, as shown in the left of Figure~\ref{fig:construct}, following the Mixture-of-Agents structure~\cite{moa}. It consists of $L$ layers. In intermediate layers ($l = 1, 2, \dots, L-1$), $n_l$ LLMs act as proposers $M_{l,i} \in \mathcal{P}$, where $\mathcal{P}$ is a pool of $N$ available models, $i=1,2,...,n_l\leq N$. Each $M_{l,i}$ processes input $x_l$ and generates a response:
\begin{equation}
\label{eq:proposer}
y_{l,i}=M_{l,i}(x_l).
\end{equation}
The output of layer $l$ is:
\begin{equation}
\label{eq:moa}
o_l = \oplus_{i=1}^{n_l} y_{l,i} + x_1, \quad x_{l+1} = o_l,
\end{equation}
where $+$ denotes concatenation and $\oplus$ denotes an aggregation prompt (see Appendix~\ref{appendix:inference_prompt}). $x_1$ is the user query. The final layer $L$ uses a single LLM to aggregate prior outputs into the final response.

To balance performance and efficiency, RouteMoA dynamically selects models for each layer $l=1,2,...,L$ through the following process:

\textbf{Step 1: Score Acquisition}. For the first layer ($l=1$), an \textit{\textbf{SLM-based scorer}} $\mathcal{S}$ performs an initial screening by predicting coarse-grained performance scores for each model in $\mathcal{P}$ on query $x_1$:
\begin{equation}
\label{eq:router}
\mathbf{s}_1 = \mathcal{S}(x_1), \quad \mathbf{s}_1 \in [0,1]^N.
\end{equation}
This scorer is not required to provide precise performance score estimates; rather, its goal is to efficiently narrow down the candidate set to a small group of high-potential models. For subsequent layers ($l>1$), a \textit{\textbf{mixture of judges}} $\mathcal{J}$ refines the initial scores by incorporating both $\mathbf{s}_1$ and responses from the previous layer, enabling more accurate and context-aware model selection:
\begin{equation}
\label{eq:adjuster}
\mathbf{s}_l = \mathcal{J}(\mathbf{s}_1, y_{l-1,1}, \dots, y_{l-1,n_{l-1}}), \quad l>1.
\end{equation}

\textbf{Step 2: Model Ranking and Selection}. The \textit{\textbf{model ranking}} module $\mathcal{R}$ selects active models for layer $l$ based on performance $\mathbf{s}_l$, cost, and latency:
\begin{equation}
\label{eq:selector}
[M_{l,1}, \dots, M_{l,n_l}] = \mathcal{R}(\mathbf{s}_l, cost, latency).
\end{equation}

An early-stopping mechanism determines when to terminate, ensuring efficient inference.

\vspace{-5pt}
\subsection{SLM-based Scorer}\label{sec:scorer}

The scorer conducts an initial screening by predicting coarse-grained performance scores to each model in the pool $\mathcal{P} = {M_1, M_2, \dots, M_N}$ given the input $\mathbf{x}_l$, as defined in Equation~\ref{eq:router}.

\noindent\textbf{Dataset generation.}
We construct a training dataset of the form:
\begin{equation}
\label{eq:d}
\mathcal{D} = \{(x^{(k)}, s^{(k)}_1,s^{(k)}_2,...,s^{(k)}_N)\}_{k=1}^{|\mathcal{D}|},
\end{equation}
where $s^{(k)}_j$ denotes the performance score of model $M_j$ on input $x^{(k)}$.
To build $\mathcal{D}$, we collect queries and ground-truth answers $\mathcal{D}_{raw}=\{(x^{(k)}, \hat{y}^{(k)})\}_{k=1}^{|\mathcal{D}|}$ from multiple datasets spanning mathematics, reasoning, coding, reading comprehension, biomedical domains, etc. For each query, we gather responses from all models in $\mathcal{P}$:
\begin{equation}
\label{eq:d1}
\mathcal{D}_{y} = \{(x^{(k)},y^{(k)}_1,y^{(k)}_2,...,y^{(k)}_N)\}_{k=1}^{|\mathcal{D}|}.
\end{equation}

Each response is scored using a combination of ground-truth accuracy and a reward model $R$ (e.g., InternLM2-1.8B-Reward):
\begin{equation}
\begin{aligned}
s^{(k)}_j &= \lambda \cdot \mathbf{1}(\hat{y}^{(k)} = y^{(k)}_j) \\
&+ (1 - \lambda) \cdot R(x^{(k)},\hat{y}^{(k)},y^{(k)}_j),
\end{aligned}
\end{equation}
where $\lambda \in (0,1)$. $\mathbf{1}(\cdot)$ is the indicator function that returns 1 if the condition is true and 0 otherwise.

\noindent\textbf{Model structure.}
Inspired by matrix factorization techniques in recommendation systems~\cite{routerdc,mf,routellm}, we model the scorer as an embedding-based similarity function.
Each model $M_j$ is assigned a learnable embedding $\mathbf{k}_j \in \mathbb{R}^d$. The input $x$ is encoded by a small language model mDeBERTaV3-base~\cite{debertav3} into an embedding $\mathcal{E}(x)$. The performance score is computed as: 
\begin{equation}
s=f(x,M_j)=\sigma(\mathcal{E}(x)^\top \mathbf{k}_j), \quad s\in[0,1],
\end{equation}
where $\sigma(\cdot)$ is the sigmoid function. Thus the full score vector is:
\begin{equation}
\label{eq:router_score}
\mathbf{s}=\mathcal{S}(x)=[f(x,M_1),f(x,M_2),...,f(x,M_N)].
\end{equation}

\noindent\textbf{Training and inference.}
During training, we adopt dual contrastive loss functions from \cite{routerdc}. The \textit{sample-LLM contrastive loss} ensures that the embeddings of models capable of answering a query are closer to the query's embedding:
\begin{flalign}
\begin{split}
 & \mathcal{L}_{\text{sample-LLM}}(x, \mathbf{s}; \theta) \\
& = \sum_{j_+ \in \mathcal{I}^+} - \log \frac{e^{\mathcal{E}(x)^\top \mathbf{k}_{j_+}}}{e^{\mathcal{E}(x)^\top \mathbf{k}_{j_+}} + \sum_{j_- \in \mathcal{I}_i^-} e^{\mathcal{E}(x)^\top \mathbf{k}_{j_-}}}, 
\end{split}
\end{flalign}
where $\mathcal{I}^+$ and $\mathcal{I}^-$ denote the top-$K_+$ and bottom-$K_-$ scoring models, respectively. $\theta$ denotes the parameters to be optimized.

The \textit{sample-sample contrastive loss} encourages semantically similar queries to have closer embeddings. 
It is formulated as:
\begin{flalign}
\begin{split}
& \mathcal{L}_{\text{sample-sample}}(x;\theta)\\
& = -\log \frac{e^{\mathcal{E}(x)^\top\mathcal{E}(x^+)}}{e^{\mathcal{E}(x)^\top\mathcal{E}(x^+)} + \sum_{x_i^- \in \mathcal{X}_i^-} e^{\mathcal{E}(x)^\top\mathcal{E}(x^-)}}.
\end{split}
\end{flalign}
where $x^+$ is a query from the same cluster as $x$, and $\mathcal{X}^-$ contains out-cluster queries. The clustering method is detailed in the Appendix~\ref{appendix:cluster}.

The total loss is:
\begin{equation}
\mathcal{L} = \mathcal{L}_{\text{sample-LLM}} + \alpha \mathcal{L}_{\text{sample-sample}},
\end{equation}
where $\alpha \geq 0$. 

\subsection{Mixture of Judges}
\label{sec:moj}
The design of the mixture of judges is motivated by two key capabilities of large language models: 
\begin{itemize}
\item \textit{Self-knowledge awareness}: Research has shown that LLMs possess the ability to evaluate their own knowledge and determine whether they understand a question~\cite{llmknow}. 
\item \textit{Cross-model evaluation}: LLMs can effectively judge responses from other models~\cite{llmjudge}, making them capable evaluators in multi-agent settings.
\end{itemize}

Thus, for layer $l>1$, we introduce mixture of judges to refine the scorer’s predictions using self- and cross-assessment signals from previous layers.

For \textbf{\textit{self-assessment}}, each active model in layer $l-1$ outputs a confidence score $s^{\text{self}}_{l-1,j}$ along with its response:
\begin{equation}
\mathbf{s}_{l-1}^{\text{self}}=[s_{l-1,1}^{\text{self}},s_{l-1,2}^{\text{self}},...,s_{l-1,n_{l-1}}^{\text{self}}].
\end{equation}

For \textbf{\textit{cross-assessment}}, to avoid the computational cost of having all models generate evaluation scores, we selectively employ only the highest-scoring model from layer $l-1$ to evaluate responses from layer $l-2$, producing scores $s^{\text{cross}}_{l-2,j}$:

\begin{equation}
\mathbf{s}_{l-2}^{\text{cross}}=[s_{l-2,1}^{\text{cross}},s_{l-2,2}^{\text{cross}},...,s_{l-2,n_{l-2}}^{\text{cross}}].
\end{equation}

Since cross-assessment relies on evaluating outputs from a prior layer, it is only applicable from the second layer onward ($l \geq 2$), as no prior outputs exist for the first layer.

The final mixture of judges function is:
\begin{align}
\mathbf{s}_l
& =\mathcal{J}(\mathbf{s}_1,y_{l-1,1}, y_{l-1,2},...,y_{l-1,n_{l-1}})\\
& =
\begin{cases}
\mathcal{U}(\mathbf{s}_1,\mathbf{s}_{l-1}^{\text{self}}), & l = 2,\\
\mathcal{U}(\mathbf{s}_1,\mathbf{s}_{l-1}^{\text{self}},\mathbf{s}_{l-2}^{\text{cross}}), &  l > 2,
\end{cases}
\end{align}
where $\mathcal{U}$ performs score normalization followed by element-wise averaging.

\subsection{Model Ranking}
\label{sec:ranking}
The model ranking module $\mathcal{R}$ selects the top-$k$ models based on the adjusted scores $\mathbf{s}_l$, with the following priority: performance > output token cost > input token cost > latency. Model pricing and latency data are sourced from OpenRouter\footnote{https://openrouter.ai/}.
Early stopping criterion is set as: 
\begin{equation}
\max(s_{l,1},s_{l,2},...,s_{l,N})>s_{th}.
\end{equation}
where $s_{th}$ is a threshold score.

If the criterion is met, or the max layer number is reached, the system will enter the aggregation stage and produce the final output:
\begin{equation}
y^{final}=M_{l,agg}(x_l).
\end{equation}

%% file: sections/4_experiment.tex
\section{Experiments}\label{sec:exp}

\subsection{Experimental Setup}
\noindent\textbf{Baselines.}
We focus on improving the computational efficiency of multi-agent collaboration while maintaining the accuracy. The compared baselines include: (1)~MoA~\cite{moa}, leveraging multiple LLMs in a layered architecture, where each agent uses outputs from previous layers to enhance its response generation; (2)~SMoA~\cite{smoa}, improving the token efficiency of MoA by employing a judge model to assess and forward only the most optimal responses to the next round. (3) To explore the impact of self-assessment and cross-assessment, we also compare RouteMoA with the version that without self-assessment and without cross-assessment for ablation study.

\noindent\textbf{Implementation Details.}
We use OpenCompass~\cite{opencompass} for data generation and evaluation. For scorer training, we employ mDeBERTaV3-base~\cite{debertav3} as the encoder, a small language model with only 86M parameters. Each LLM embedding is projected to a 768-dimensional vector space. The training parameters are set as $\alpha=0.2$ and $\lambda=0.5$, which are observed to be insensitive within the ranges of $[0.2,2]$ and $[0.3,0.9]$, respectively. The number of k-means clusters is set to $6$. Training is conducted using the AdamW optimizer with a learning rate of $5 \times 10^{-5}$, a weight decay of $0.01$, and a mini-batch size of 64. We report average performance, cost, and latency. Experiments are run on 80GB GPUs. 

\input{tex_files/large_acc}

\input{tex_files/small_acc}
\input{tex_files/ood}

\noindent\textbf{Exp1: Scalability Evaluation on Large-Scale Model Pool.} To validate the practical scalability and efficiency of RouteMoA in real-world deployment scenarios, we construct a large-scale model pool consisting of \textit{\textbf{15 state-of-the-art LLMs}} of varying sizes (from 4B to 235B parameters) and capabilities, including general-purpose, reasoning-specialized, and code/math-focused models (see Table~\ref{tab:large_model_pool} in Appendix). Notably, this pool contains models with both standard (\textit{no-think}) and advanced reasoning (\textit{think}) modes, presenting a diverse and challenging testbed for multi-agent collaboration. 

\textbf{Evaluation Benchmark.} We conduct a comprehensive evaluation on a collection of \textit{\textbf{30 datasets}} spanning five critical capability categories: Language Understanding, Reading \& QA, Logic Reasoning, Math Reasoning, and Language Generation (see Table~\ref{tab:large_data_test},~\ref{tab:large_data_train} for the full list). This broad coverage ensures a rigorous assessment of generalizability.

\noindent\textbf{Exp2: Performance on Small-Scale Model Pool.} To enable a direct and fair comparison with MoA and SMoA (which are limited to small pools due to their full-model inference design), we further evaluate on a compact but diverse pool of 5 LLMs: Gemma-2-9B-it~\cite{gemma}, Ministral-8B-Instruct~\cite{mistral}, Qwen2.5-Coder-7B-Instruct~\cite{qwencoder}, Qwen2.5-Math-7B-Instruct~\cite{qwenmath}, and Bio-Medical-Llama-3-8B~\cite{biomed}. The evaluation covers 5 datasets (MATH-500~\cite{math}, ARC-Challenge~\cite{arc}, MBPP~\cite{mbpp}, RACE-high~\cite{race}, MMLU-bio~\cite{mmlu}) representing mathematics, reasoning, coding, reading, and biomedical knowledge.

\noindent\textbf{Exp3: Out-of-Distribution Generalization.} We further evaluate generalization on the challenging AGIEval-Gaokao~\cite{agieval} benchmark, which spans nine subjects (Biology, Chemistry, Chinese, English, Geography, History, MathCloze, MathQA, Physics). This human-exam benchmark tests the model's ability to handle diverse, unseen tasks requiring human-like reasoning.



\begin{table}[t]

\centering
\caption{\textbf{Ablation study} for mixture of judges on small-scale model pool.}
\vspace{-5pt}
\label{tab:ablation}
\resizebox{0.47\textwidth}{!}{
\begin{tabular}{l|ccc}
\toprule
Method & Performance(\%) $\uparrow$ & Cost(\$) $\downarrow$ & Latency(s) $\downarrow$ \\
\midrule
RouteMoA     & \textbf{83.1} & 7.68 & 10.64 \\
w/o self. & 82.6 & 7.99 & 10.49  \\
w/o cross.  & 82.7 & \textbf{7.25} & \textbf{10.29}  \\
\bottomrule
\end{tabular}
}
\vspace{-13pt}
\end{table}

\subsection{Main Results}
\noindent\textbf{Scalability Evaluation on Large-Scale Model Pool.}
As shown in Table~\ref{tab:large_scale}, RouteMoA demonstrates exceptional scalability, performance, and efficiency. It achieves an average accuracy of 78.6, significantly surpassing MoA (71.3) and SMoA (69.7), with especially large gains in Math Reasoning (+47.5\%) and Language Generation (+23.9\%). Unlike MoA and SMoA, which lack a clear model selection criteria and become infeasible at scale due to prohibitive costs and context limits, RouteMoA remains practical by dynamically routing queries to an optimal model subset. This approach reduces total cost by 89.8\% and latency by 63.6\% compared to MoA, while also outperforming SMoA in both efficiency and accuracy.

Furthermore, RouteMoA consistently achieves the best accuracy, lowest cost, and lowest latency across all five capability categories. In Language Understanding and Reading\&QA, it matches or surpasses MoA’s accuracy while reducing cost by 95.3\%. These results demonstrate that dynamic routing tailored for multi-agent systems enables efficient and effective collaboration in large, heterogeneous model pools, particularly for complex tasks requiring complementary model strengths.

\noindent\textbf{Performance on Small-Scale Model Pool.}
As shown in Table~\ref{tab:performance}, RouteMoA substantially improves efficiency, reducing inference cost by 81.4\% compared to MoA (6.71 vs. 36.03), and by 88.2\% on domain-specific scenarios such as MMLU-bio, demonstrating effective avoidance of expensive generalist models. It also lowers average latency by 38.7\% (10.01s vs. 16.32s) due to lightweight scoring and targeted model selection. Meanwhile, RouteMoA achieves the highest average score (83.1), outperforming single models and MoA, with statistically significant gains over SMoA (paired t-test shows t = 2.296, p = 0.0217 < 0.05). These results confirm that routing combined with response aggregation forms an effective paradigm for multi-LLM collaboration.

\noindent\textbf{Out-of-Distribution Generalization.}
As shown in Table~\ref{tab:comparison}, RouteMoA outperforms SMoA with higher average accuracy (54.62 vs. 52.92) while reducing cost by 11.5\% and latency by 24.7\%. It achieves notable accuracy gains in humanities and science subjects, including Geography (+7.04), History (+5.10), Physics (+4.50), and Biology (+4.77). These results demonstrate that RouteMoA effectively exploits specialized models on unseen tasks requiring human-like reasoning, exhibiting strong out-of-distribution generalization.

\begin{figure}[t]
    \centering
    \includegraphics[width=0.43\textwidth]{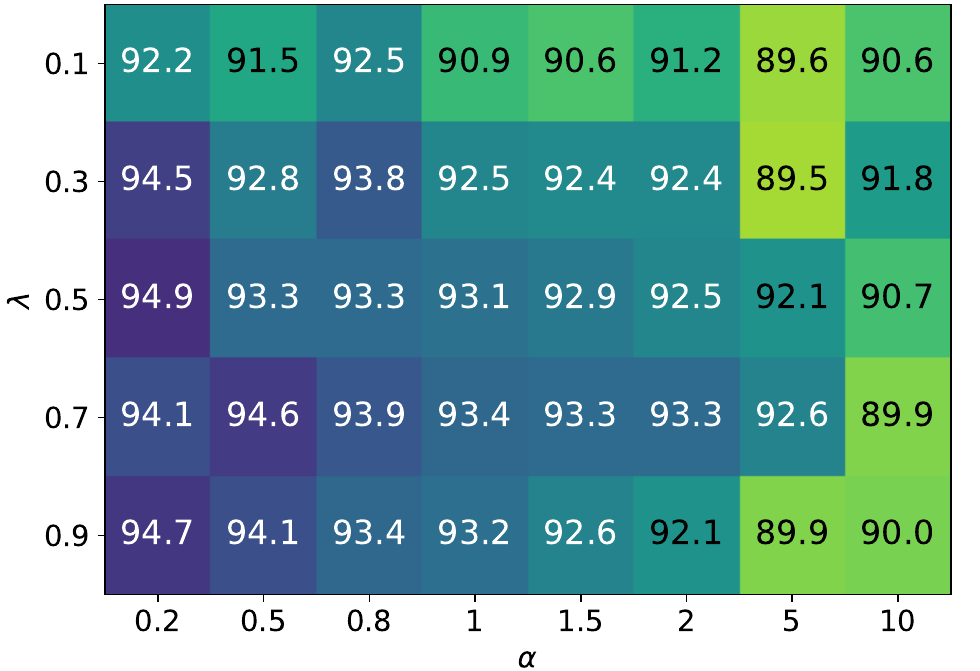}
    \vspace{-5pt}
    \caption{
    \textbf{Average values of three scorer assessment metrics} (Top-1-Hit, Top-3-Hit, and Top-3-Agree) under different training hyperparameters ($\lambda$ and $\alpha$).}
    \label{fig:alpha_lambda}
    \vspace{-10pt}
\end{figure}

\begin{figure*}[t]
    \centering
    \includegraphics[width=0.99\textwidth]{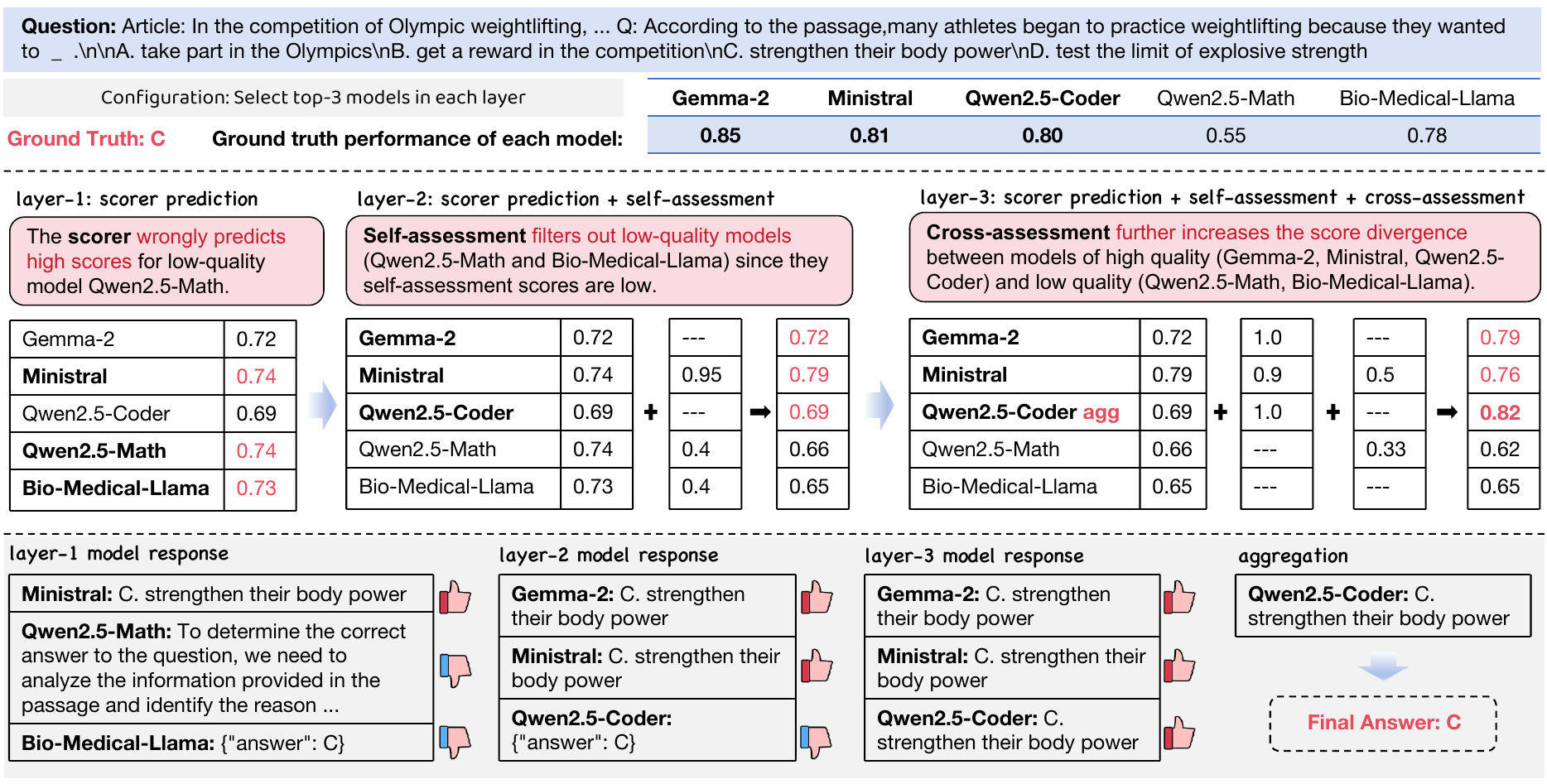}
    \vspace{-5pt}
    \caption{\textbf{Case study} of adjusting wrong scorer predictions with self- and cross-assessment.} 
    \label{fig:example}
    \vspace{-10pt}
\end{figure*}

\subsection{Analysis}

\textbf{Scorer Assessment.} 
In RouteMoA, the scorer plays a key role in providing an initial screening of candidate models. Rather than requiring precise performance prediction, the scorer is designed to identify a small set of high-potential models for subsequent refinement via mixture-of-judges (including self-assessment and cross-assessment). To evaluate its effectiveness, we introduce three metrics: Top-1 Hit Rate (Top-1-Hit), Top-3 Hit Rate (Top-3-Hit) and Top-3 Agreement Rate (Top-3-Agree). Top-1-Hit and Top-3-Hit measure the probability that the ground-truth best model appears in the scorer’s top-one / top-three predictions, which reaches 90.7\% and 97.9\% when $\alpha=0.2, \lambda=0.5$. The Top-3-Agree quantifies the overlap between the scorer’s top-three selections and the ground-truth top-three models, achieving 96.2\%, indicating that the scorer successfully narrows down the candidate pool to a small subset containing high-performing models in the majority of cases. The detailed calculation method and results under different $\alpha, \lambda$ combinations are shown in Appendix~\ref{appendix:scorer_metric}.


\noindent\textbf{Mixture-of-Judges Ablation Study.}
To explore the impact of self-assessment and cross-assessment in mixture of judges, we conduct ablation studies. As shown in Table~\ref{tab:ablation}, the average performance of RouteMoA without self-assessment is 82.6, and the performance without cross-assessment is 82.7, both of them are lower than the 83.1 achieved by RouteMoA. RouteMoA without cross-assessment achieves the lowest cost and latency. It is a natural result since the additional judging token will not be generated without cross-assessment.

\noindent\textbf{Case Study.} To illustrate how the routing pipeline, especially the mixture of judges works, we present an example from the RACE-high dataset in Figure~\ref{fig:example}. In this dataset, Gemma-2, Ministral, and Qwen2.5-Coder usually perform well, whereas Qwen2.5-Math and Bio-Medical-Llama show weaker performance. 
In layer-1, the scorer incorrectly assigns a high score (0.74) to Qwen2.5-Math, while giving a relatively low score to the high-quality model Qwen2.5-Coder. In layer-2, models generate self-assessment scores. Qwen2.5-Math and Bio-Medical-Llama produce low self-scores due to their inability to accurately follow instructions, thus the self-assessment mechanism effectively filters out low-quality models. In layer-3, cross-assessment further widens the score gap between high- and low-quality models, since a low cross-score is assigned to Qwen2.5-Math. Examining the model responses across layers, we observe increasing participation of high-quality models and progressive improvement in response quality as the number of layers increases. Eventually, models achieve consensus at the final layer. 
This correction process is supported by the high Top-3 Hit Rate (97.9\%) of the scorer, indicating that in 97.9\% of cases, at least one correct model is included in the initial candidate set. Once present, the multi-agent collaboration mechanism effectively identifies and amplifies the correct response through answer aggregation and self/cross-assessment.

%% file: tex_files/large_acc.tex
\newcommand{\good}[1]{{\fontsize{7}{11}\selectfont\textcolor[HTML]{068c73}{$\uparrow$#1\%}}}
\newcommand{\bad}[1]{{\fontsize{7}{11}\selectfont\textcolor[HTML]{f42c55}{$\downarrow$#1\%}}}

\begin{table*}[h]
\setlength\tabcolsep{3pt}
\centering
\footnotesize
\caption{
\textbf{Performance and efficiency comparison on the large-scale model pool (15 LLMs).}
\textcolor[HTML]{068c73}{$\uparrow$} indicates an improvement over MoA, while \textcolor[HTML]{f42c55}{$\downarrow$} represents a degradation compared to MoA. Both are denoted by percentage.
}
\vspace{-5pt}
\label{tab:large_scale}
\resizebox{\textwidth}{!}{
\begin{tabular}{l|l|ccccc|c}
\toprule
 & Method & Language Understanding & Reading\&QA & Logic Reasoning & Math Reasoning & Language Generation & Avg. \\
\midrule
\multirow{3}{*}{\textbf{Accuracy (\%)} $\uparrow$} 
 & MoA & 83.4  & 88.0 & 93.3 & 49.7 & 41.9 & 71.3 \\
 & SMoA & 78.4 \bad{6.00}  & 85.3 \bad{3.07} & 91.1 \bad{2.36} & 53.1 \good{6.84} & 40.3 \bad{3.82} & 69.7 \bad{2.24} \\
 & \cellcolor{tableblue}RouteMoA & \cellcolor{tableblue}\textbf{84.0} \good{0.72}  & \cellcolor{tableblue}\textbf{88.0} \good{0.00} & \cellcolor{tableblue}\textbf{95.6} \good{2.50} & \cellcolor{tableblue}\textbf{73.3} \good{47.5} & \cellcolor{tableblue}\textbf{51.9} \good{23.9} & \cellcolor{tableblue}\textbf{78.6} \good{10.2} \\
\midrule
\multirow{3}{*}{\textbf{Cost (\$)} $\downarrow$} 
 & MoA & 321.7  & 303.4 & 385.3 & 751.1 & 477.5 & 447.8 \\
 & SMoA & 47.8 \good{85.1}  & 53.8 \good{82.3} & 57.2 \good{85.2} & 232.5 \good{69.0} & 110.4 \good{76.9} & 100.4 \good{77.6} \\
 & \cellcolor{tableblue}RouteMoA & \cellcolor{tableblue}\textbf{24.9} \good{92.3}  & \cellcolor{tableblue}\textbf{14.2} \good{95.3} & \cellcolor{tableblue}\textbf{28.7} \good{92.6} & \cellcolor{tableblue}\textbf{94.6} \good{87.4} & \cellcolor{tableblue}\textbf{65.7} \good{86.2} & \cellcolor{tableblue}\textbf{45.6} \good{89.8} \\
\midrule
\multirow{3}{*}{\textbf{Latency (s)} $\downarrow$}
 & MoA & 126.3  & 101.4 & 134.2 & 619.5 & 258.9 & 248.1 \\
 & SMoA & 76.4 \good{39.5}  & 94.1 \good{7.20} & 76.1 \good{43.3} & 471.2 \good{23.9} & 257.1 \good{0.70} & 195.0 \good{21.4} \\
 & \cellcolor{tableblue}RouteMoA & \cellcolor{tableblue}\textbf{43.4} \good{65.6}  & \cellcolor{tableblue}\textbf{27.8} \good{72.6} & \cellcolor{tableblue}\textbf{58.9} \good{56.1} & \cellcolor{tableblue}\textbf{211.4} \good{65.9} & \cellcolor{tableblue}\textbf{109.3} \good{57.8} & \cellcolor{tableblue}\textbf{90.2} \good{63.6} \\
\bottomrule
\end{tabular}
}
\vspace{-5pt}
\end{table*}

%% file: tex_files/small_acc.tex
\begin{table*}[t]
\setlength\tabcolsep{7pt}
\centering
\footnotesize
\caption{
\textbf{Performance and efficiency comparison on the small-scale model pool (5 LLMs)}. \textit{Oracle} means using ground truth assessment scores for LLM selection. Best results of multi-LLM methods are bold. A paired t-test confirms that the improvement of RouteMoA over SMoA is statistically significant (t = 2.296, p = 0.0217 < 0.05).
}
\vspace{-5pt}
\label{tab:performance}
\resizebox{\textwidth}{!}{
\begin{tabular}{l|l|ccccc|c}
\toprule
 \multicolumn{8}{c}{\textbf{Accuracy(\%)} $\uparrow$} \\
\midrule
 Type & Method & MATH & ARC-c & MBPP & RACE-high & MMLU-bio & Avg. \\
 \midrule

\multirow{5}{*}{Single LLM} 
 & Gemma-2-9B-it & 46.5  & 90.2 & 66.2 & 85.6 & 78.6 & 75.9 \\
 & Ministral-8B-Instruct-2410 & 51.0  & 85.3 & 63.0 & 80.3 & 70.7 & 72.7 \\
 & Qwen2.5-Coder-7B-Instruct & 65.3  & 85.5 & 79.8 & 80.6 & 67.2 & 77.5 \\
 & Qwen2.5-Math-7B-Instruct & 80.7  & 50.6 & 52.9 & 55.0 & 43.5 & 63.0 \\
 & Bio-Medical-Llama-3-8B & 11.7 & 75.1 & 16.0 & 77.3 & 87.0 & 46.2 \\
\midrule
\multirow{3}{*}{Single LLM with Routing} 
& \textcolor{gray}{Oracle} 
& \textcolor{gray}{83.8} 
& \textcolor{gray}{96.8} 
& \textcolor{gray}{86.7} 
& \textcolor{gray}{94.5} 
& \textcolor{gray}{95.6} 
& \textcolor{gray}{92.5} \\
 & RouteLLM & 64.3  & 84.8 & 76.3 & 79.4 & 65.7 & 76.2 \\
 & RouterDC & 72.8  & 87.3 & 72.6 & 78.3 & 70.9 & 78.9 \\
\midrule
\multirow{2}{*}{Multi-LLMs}
& MoA & 73.6  & 87.0 & 75.5 & 80.1 & 76.0 & 80.9 \\
& SMoA & 73.5 \bad{0.10}  & \textbf{89.4} \good{2.80} & 79.4 \good{5.20} & \textbf{84.0} \good{4.90} & 75.7 \bad{0.40} & 82.6 \good{2.10} \\
\midrule
\multirow{1}{*}{\textbf{Ours}}
& 
\cellcolor{tableblue}RouteMoA 
& \cellcolor{tableblue}\textbf{76.0} \good{3.30} 
& \cellcolor{tableblue}88.2 \good{1.40}
& \cellcolor{tableblue}\textbf{79.8} \good{5.70}
& \cellcolor{tableblue}81.0 \good{1.10}
& \cellcolor{tableblue}\textbf{79.3} \good{4.30}
& \cellcolor{tableblue}\textbf{83.1} \good{2.70} \\

\bottomrule
\end{tabular}
}
\setlength\tabcolsep{1pt}
\resizebox{\textwidth}{!}{
\begin{tabular}{l|cccccc|cccccc}
\toprule
Resource
 & \multicolumn{6}{c|}{\textbf{Cost (\$)} $\downarrow$} 
 & \multicolumn{6}{c}{\textbf{Latency (s)} $\downarrow$} \\
 \midrule
Dataset 
& MATH & ARC-c & MBPP & RACE-high & MMLU-bio & Total 
& MATH & ARC-c & MBPP & RACE-high & MMLU-bio & Avg. \\
\midrule
MoA
& 19.68 & 2.27 & 0.61 & 8.53 & 1.78 & 36.03
& 26.62 & 12.05 & 15.52 & 13.45 & 14.07 & 16.32 \\
SMoA
& 4.40\good{77.6} & 0.47\good{79.3} & \textbf{0.18}\good{70.5} & 2.22\good{74.0} & 0.36\good{79.8} & 8.23\good{77.2}
& 23.16\good{13.0} & 10.45\good{13.3} & 10.30\good{33.6} & 11.02\good{18.1} & 11.86\good{15.7} & 13.31\good{18.4} \\
\midrule
\rowcolor{tableblue}
RouteMoA
& \textbf{4.03}\good{79.5} & \textbf{0.28}\good{87.7} & 0.37\good{39.3} & \textbf{1.82}\good{78.7} & \textbf{0.21}\good{88.2} & \textbf{6.71}\good{81.4}
& \textbf{19.05}\good{28.4} & \textbf{9.73}\good{19.3} & \textbf{7.31}\good{52.9} & \textbf{4.45}\good{66.9} & \textbf{9.51}\good{32.4} & \textbf{10.01}\good{38.7} \\
\bottomrule
\end{tabular}
}
\label{tab:cost_latency}
\vspace{-10pt}
\end{table*}

%% file: tex_files/ood.tex
\begin{table*}[t]
\centering
\setlength\tabcolsep{3pt}
\vspace{10pt}
\caption{\textbf{Out-of-distribution benchmark comparison} between SMoA and RouteMoA.}
\vspace{-5pt}
\label{tab:comparison}
\resizebox{\textwidth}{!}{
\begin{tabular}{l|l|ccccccccc|c}
\toprule
& Method & Biology & Chemistry & Chinese & English & Geography & History & MathCloze & MathQA & Physics & OOD Avg. \\
\midrule
\multirow{2}{*}{\textbf{Accuracy (\%)} $\uparrow$} &
SMoA     & 53.33 & 37.68 & 49.59 & 80.39 & 60.80 & 64.26 & 27.12 & 64.10 & 39.00 & 52.92 \\
&\cellcolor{tableblue}RouteMoA & \cellcolor{tableblue}58.10 & \cellcolor{tableblue}37.68 & \cellcolor{tableblue}49.19 & \cellcolor{tableblue}77.78 & \cellcolor{tableblue}67.84 & \cellcolor{tableblue}69.36 & \cellcolor{tableblue}27.12 & \cellcolor{tableblue}60.97 & \cellcolor{tableblue}43.50 & \cellcolor{tableblue}\textbf{54.62} \\
\midrule
\multirow{2}{*}{\textbf{Cost} (\$) $\downarrow$} &
SMoA     & 4.71 & 6.90 & 8.05 & 5.25 & 3.80 & 3.49 & 8.79 & 9.04 & 7.31 & 6.37 \\
&\cellcolor{tableblue}RouteMoA & \cellcolor{tableblue}4.18 & \cellcolor{tableblue}7.77 & \cellcolor{tableblue}6.33 & \cellcolor{tableblue}3.46 & \cellcolor{tableblue}4.15 & \cellcolor{tableblue}3.33 & \cellcolor{tableblue}6.08 & \cellcolor{tableblue}7.57 & \cellcolor{tableblue}7.91 & \cellcolor{tableblue}\textbf{5.64} \\
\midrule
\multirow{2}{*}{\textbf{Latency} (s) $\downarrow$} &
SMoA     & 11.25 & 15.41 & 10.90 & 9.40 & 9.79 & 9.86 & 19.09 & 19.85 & 15.91 & 13.50 \\
&\cellcolor{tableblue}RouteMoA & \cellcolor{tableblue}6.12 & \cellcolor{tableblue}14.46 & \cellcolor{tableblue}4.57 & \cellcolor{tableblue}2.71 & \cellcolor{tableblue}5.59 & \cellcolor{tableblue}4.26 & \cellcolor{tableblue}17.72 & \cellcolor{tableblue}20.08 & \cellcolor{tableblue}15.93 & \cellcolor{tableblue}\textbf{10.16} \\
\bottomrule
\end{tabular}
}
\vspace{-10pt}
\end{table*}



%% file: sections/5_conclusion.tex
\section{Conclusion}\label{sec:conclusion}

In conclusion, we present RouteMoA, an efficient Mixture-of-Agents framework that overcomes the resource limitations of classical MoA through dynamic routing. The framework employs a lightweight scorer to perform an initial screening of candidates using prior knowledge from the query, followed by a mixture of judges that refines scores with posterior knowledge from model outputs. RouteMoA significantly reduces cost and latency while maintaining strong performance. Experimental results also show strong OOD generalization ability and large-scale model pool scalability. This prior-posterior routing approach offers a scalable and practical path toward efficient multi-LLM collaboration.


\section{Limitation}
The scorer requires retraining to support new LLMs. However, integrating a new LLM only involves training a lightweight scorer on a small curated query set, which takes about 25 minutes. Future work will explore retrain-free routing.

%% file: sections/7_appendix.tex
\appendix
\section{Aggregation Prompt, Self- and Cross-assessment Prompt}
\label{appendix:inference_prompt}
The prompts used in the inference stage of RouteMoA are shown in Figure~\ref{prompt:layer1}, Figure~\ref{prompt:layer2}, and Figure~\ref{prompt:layer3}. These prompts include instructions for the LLM to aggregate responses from models in the previous layer, to score its own answer (self-assessment prompt), and to evaluate answers from other LLMs (cross-assessment prompt). Specifically, the prompt at layer-1 (Figure~\ref{prompt:layer1}) includes both the aggregation prompt and the self-assessment prompt. For intermediate layer-$l|_{1<l<L}$ (Figure~\ref{prompt:layer2}), the prompts consist of the aggregation prompt, the self-assessment prompt, and the cross-assessment prompt. Finally, in the last layer (Figure~\ref{prompt:layer3}), only the aggregation prompt is included.

\input{tex_files/layer1_prompt}
\input{tex_files/layer2_prompt}
\input{tex_files/layer3_prompt}

\section{Prompt for Training Dataset Generation}

\label{appendix:datagen_prompt}
To generate training data that meets the requirements of layer-1, we prompt the LLMs to answer questions from the datasets. The prompts used for each dataset are presented in Figure~\ref{prompt:single}. The model outputs are then compared with the ground truth answers. For layer-$l|_{l>1}$, we prompt the models to generate aggregated answers based on reference answers from all models. These aggregated responses are subsequently evaluated against the ground truth. Additionally, a judge model is employed to assess the quality of each answer. The prompt used for generating the aggregated answers is shown in Figure~\ref{prompt:data_gen}. We use InternLM2-1.8B-Reward as the judge model.

\input{tex_files/data_gen_prompt}
\input{tex_files/single_prompt}

\section{Clustering Details for Sample-Sample Loss}
\label{appendix:cluster}
The \textit{sample-sample contrastive loss} encourages semantically similar queries to have closer embeddings. To achieve this, we use t-SNE~\cite{tsne} and k-means~\cite{kmeans} algorithm to transfer input prompt embeddings to low-dimensional vectors and cluster them into $Q$ groups$\{\mathcal{K}_1, \mathcal{K}_2,...,\mathcal{K}_Q\}$. We randomly select an in-group query $x^+\in \mathcal{K}_q$, and an out-group set $\mathcal{X}^-\subset \{\cup_{q'\neq q}\mathcal{K}_{q'}\}$ of $H$ queries from the training mini-batch.

\section{Top-1-Hit, Top-3-Hit and Top-3-Agree Calculation Details}
\label{appendix:scorer_metric}
In this section, we provide detailed definitions and calculation methods for the three evaluation metrics used to assess the effectiveness of the scorer in RouteMoA: 
Top-1 Hit Rate (Top-1-Hit), Top-3 Hit Rate (Top-3-Hit), and Top-3 Agreement Rate (Top-3-Agree).

\subsection*{Top-1 Hit Rate (Top-1-Hit)}
The Top-1 Hit Rate (Top-1-Hit) measures the probability that the scorer's top-1 prediction is one of the models that are able to provide a correct answer.

Let $p$ be the index of the top 1 model according to the scorer, and $T =\{t_1,...,t_k\}$ be the set of indices of the models that are able to provide a correct answer.

We define the Top-1 Hit Rate as:

$$
\text{Top-1-Hit} = 
\begin{cases}
1, & \text{if } p \in T, \\
0, & \text{otherwise}.
\end{cases}
$$

The final Top-1-Hit is obtained by averaging this score over all test cases that can be answered correctly by at least one model.

\subsection*{Top-3 Hit Rate (Top-3-Hit)}
The Top-3 Hit Rate (Top-3-Hit) measures the probability that the ground-truth best model is included in the scorer's top-three predictions.

Let $ P = \{p_1, p_2, p_3\} $ be the set of indices of the top 3 models according to the scorer, and $T =\{t_1,...,t_k\}$ be the set of indices of the models that are able to provide a correct answer.

We define the Top-3 Hit Rate as:

$$
\text{Top-3-Hit} = 
\begin{cases}
1, & \text{if } |P \cap T| \geq 1, \\
0, & \text{otherwise}.
\end{cases}
$$

The final Top-3-Hit is obtained by averaging this score over all test cases that can be answered correctly by at least one model.

\subsection*{Top-3 Agreement Rate (Top-3-Agree)}

The Top-3 Agreement Rate (Top-3-Agree) is a metric to evaluate whether the top three models selected by the scorer align with those that have the best true performance. For each test case or dataset, the scorer selects the top 3 models based on predicted scores, and we compare them with the top 3 models according to ground truth performance.

Let $ P = \{p_1, p_2, p_3\} $ be the set of indices of the top 3 models according to the scorer, and $ T = \{t_1, t_2, t_3\} $ be the set of indices of the top 3 models according to the ground truth.

We define the Top-3 Agreement Rate as:

$$
\text{Top-3-Agree} =
\begin{cases}
1, & \text{if } |P \cap T| = 3, \\
0.6, & \text{if } |P \cap T| = 2, \\
0.3, & \text{if } |P \cap T| = 1, \\
0, & \text{otherwise}.
\end{cases}
$$

This scoring rule assigns a full score if all top 3 models are correctly identified, a partial score if two / one out of the top 3 are correct, and zero otherwise. The final Top-3-AR is obtained by averaging this score over all test cases that can be answered correctly by at least one model.

\subsection*{Scorer evaluation under different $\alpha$ and $\lambda$}

\begin{figure}[htbp]
    \centering
    \includegraphics[width=0.43\textwidth]{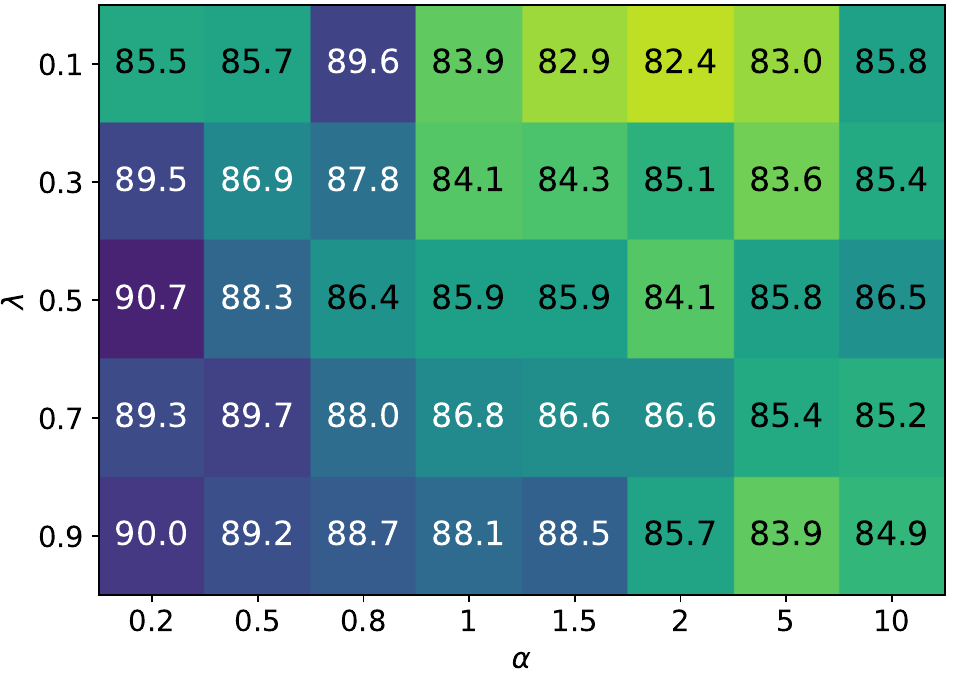}
    \vspace{-5pt}
    \caption{
    Top-1-Hit under different training hyperparameters ($\lambda$ and $\alpha$) for the scorer module.}
    \label{fig:top_1_hit}
    \vspace{-10pt}
\end{figure}

\begin{figure}[htbp]
    \centering
    \includegraphics[width=0.43\textwidth]{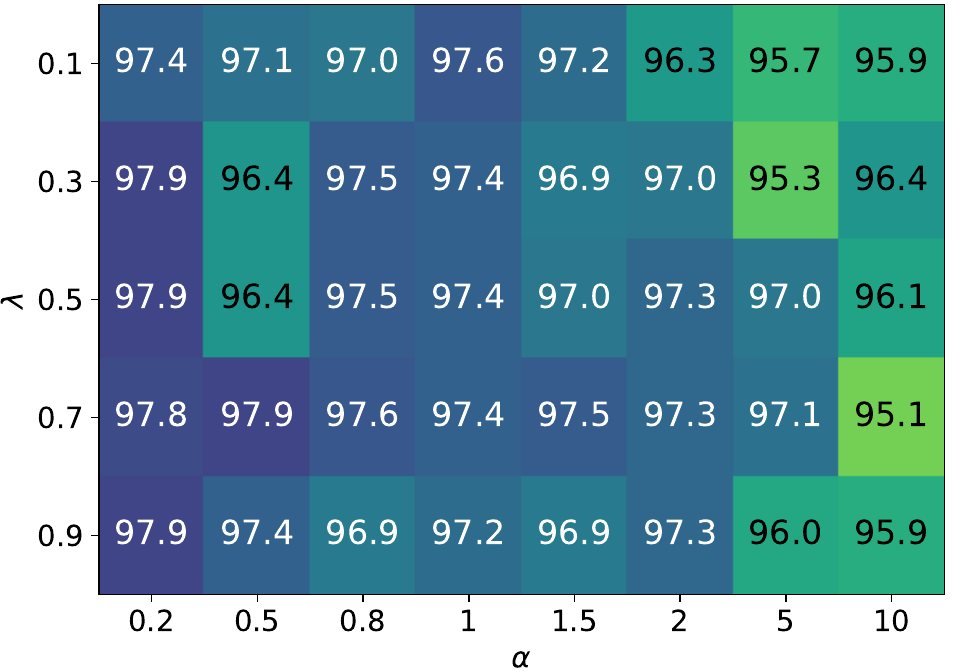}
    \vspace{-5pt}
    \caption{
    Top-3-Hit under different training hyperparameters ($\lambda$ and $\alpha$) for the scorer module.}
    \label{fig:top_3_hit}
    \vspace{-10pt}
\end{figure}

\begin{figure}[htbp]
    \centering
    \includegraphics[width=0.43\textwidth]{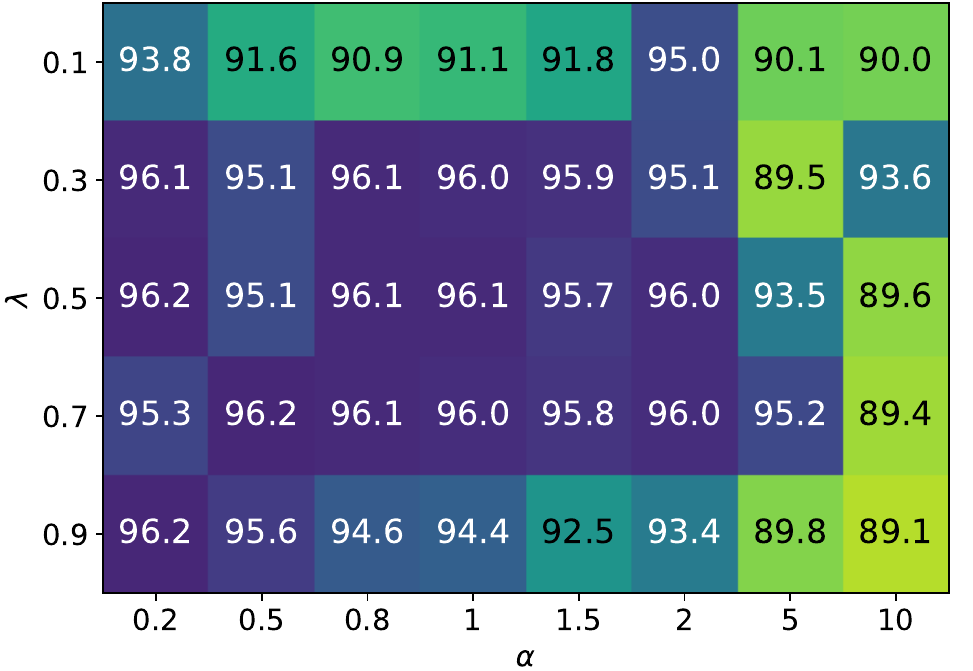}
    \vspace{-5pt}
    \caption{
    Top-3-Agree under different training hyperparameters ($\lambda$ and $\alpha$) for the scorer module.}
    \label{fig:top_3_ar}
    \vspace{-10pt}
\end{figure}

We calculate the three evaluation metrics: Top-1-Hit, Top-3-Hit, and Top-3-Agree under different combinations of scorer training hyperparameters $\alpha$ and $\lambda$, the results are shown in Figure~\ref{fig:top_1_hit}, \ref{fig:top_3_hit}, \ref{fig:top_3_ar}, respectively.

\section{Dataset Statistics for Scorer Training of Small-Scale Model Pool}

The dataset statistics for scorer training are presented in Table~\ref{tab:stat}. The train/dev/test splits generally follow the original partitioning of each dataset. For datasets that lack a dev split, we further divide the original training portion into train and dev subsets. 

\begin{table}[htbp]
\setlength\tabcolsep{7pt}
\centering
\footnotesize
\centering
\begin{tabular}{l|ccc}
\toprule
Dataset & train & dev & test \\
\midrule
 MATH & 7125 & 375 & 5000 \\
 ARC-c & 1119 & 299 & 1165 \\
 MBPP & 120 & 43 & 257 \\
RACE-high & 7000 & 300 & 3498 \\
 MMLU-biomed & 92 & 25 & 817\\
\bottomrule
\end{tabular}
\caption{Dataset statistics for scorer training.}
\label{tab:stat}
\end{table}

\section{Evaluation Details on Large-Scale Model Pool}
\label{appendix:large_scale}
To explore the ability of RouteMoA to handle large-scale agent pools, we conduct experiments with an agent pool containing 15 newest LLMs including Qwen~\cite{qwen3} and Deepseek~\cite{deepseek-r1, deepseek-v3} series with varying sizes (from 4B to 235B parameters), capabilities, and think mode, as listed in Table~\ref{tab:large_model_pool}.

Note that both MoA and SMoA are infeasible to handle the larger agent pool, since these methods need all LLMs to infer and then aggregate the responses. When the agent pool is large, it is a huge cost for all model to perform inference. Besides, the context will be too long and exceeds the limit. Although we can choose a subset of agents from the large agent pool for MoA and SMoA to alleviate these problems, these two methods lack clear criteria for selecting such model subsets. If we select high-performing models such as Deepseek-R1, the total cost will be high, and the context will be long. If we select smaller models such as Qwen2.5-7B, although the cost will be lower, the performance can not be guaranteed.

In contrast, our RouteMoA method is designed to deal with such situation. It has clear criteria on how to select agent subsets according to the categories and complexity of user queries. It lowers cost while ensures a competitive performance. Specifically, to handle the larger agent pool, we collect a wide range of datasets as a query pool, shown in Table~\ref{tab:large_data_train}. We only select a subset from each dataset to enable that the whole query pool is not very large. The dataset statistics are also shown in Table~\ref{tab:large_data_train}. Maintaining a smaller query pool benefits the scalability of the scorer. If a new LLM is added into the agent pool, it infers on these queries (the process will be shorter if the query pool is relatively small) and results are used to train a new scorer. The training process can be completed in less than 30 minutes, using less than 50GB GPU memory. 

We evaluate RouteMoA on 30 test sets, as shown in Table~\ref{tab:large_data_test}, each set contains 15 samples (not overlapping with the training set). Among these, lcqmc~\cite{lcqmc}, mrpc~\cite{mrpc}, and cluewsc2020~\cite{cluewsc} are out-of-distribution test sets, which do not appear in the training set.

\section{Model \& Data License and Intended Use Statement}
Our experiments utilize a collection of publicly available models and benchmark datasets. To the best of our knowledge, our use of these models and datasets is consistent with their intended research purposes as specified by their original creators. All models and datasets used in this work are cited. For any model or dataset we use, we adhere to its stipulated terms of use.

\begin{table}[htbp]
\setlength\tabcolsep{6pt}
\centering
\footnotesize
\setlength\tabcolsep{2.5pt}
\centering

\begin{tabular}{l|l|l}
\toprule
Size & Model & Thinking Mode\\
\midrule
\multirow{3}{*}{Small} & Qwen3-4B & no-think \\
 & Qwen3-8B & no-think \\
 & Qwen2.5-7B-Instruct & no-think \\
 \midrule
\multirow{6}{*}{Medium} & Qwen3-14B & no-think \\
 & Qwen3-32B & no-think \\
 & Qwen3-30B-A3B & think/no-think \\
 & QwQ-32B & think \\
 & DeepSeek-R1-Distill-Qwen-14B & think \\
 & DeepSeek-R1-Distill-Qwen-32B & think \\
 \midrule
\multirow{6}{*}{Large} & Qwen2.5-72B-Instruct & no-think \\
 & Qwen3-235B-A22B & think/no-think \\
 & DeepSeek-R1 & think \\
 & DeepSeek-R1-0528 & think \\
& DeepSeek-V3 & no-think \\
& DeepSeek-V3-0324 & no-think \\

\bottomrule
\end{tabular}
\caption{Models that forms a larger agent pool.}
\label{tab:large_model_pool}

\end{table}

\begin{table}[htbp]
\setlength\tabcolsep{6pt}
\centering
\footnotesize
\setlength\tabcolsep{2.5pt}
\centering

\begin{tabular}{l|l}
\toprule
Category & Dataset \\
\midrule
\multirow{4}{*}{Language Understanding} & lcqmc, ocnli, sst2  \\
 & cola, mrpc, msra, qqp \\
 & sts\_b, ag\_news \\
 & qnli, chid\_baidu \\
 \midrule
\multirow{5}{*}{Reading\&QA} & webqa \\
 & c3 \\
 & cmrc \\
 & race \\
 & story\_cloze \\
 \midrule
\multirow{3}{*}{Logic Reasoning} & cluewsc2020 \\
 & winogrande\_wsc \\
 & truthful\_qa \\
\midrule
\multirow{6}{*}{Math Reasoning} & bigmath \\
 & gsm8k \\
 & GAOKAO-2023\_Math\_en \\
 & geometry \\
& prealgebra \\
& precalculus \\
\midrule
\multirow{5}{*}{Language Generation} & word\_manipulation\_v2 \\
 & nlpcc2017\_task2 \\
 & lcsts \\
& nlpcc2018\_task2 \\
& conll2014 \\
\bottomrule
\end{tabular}
\caption{The test dataset categories for the large agent pool.}
\label{tab:large_data_test}

\end{table}

\begin{table*}[htbp]
\setlength\tabcolsep{7pt}
\centering
\footnotesize
\begin{tabular}{l|l|c}
\toprule
Dataset & Description & Data Num \\
\midrule
ag\_news~\cite{agnews} & News topic classification & 80 \\
algebra~\cite{mmlu} & Algebra math problems & 80 \\
BBH-100~\cite{bbh} & BIG-Bench Hard subset & 80 \\
bigmath~\cite{bigmath} & Complex math problems & 16 \\
c3~\cite{c3} & Chinese multiple-choice QA & 80 \\
chid~\cite{chid} & Chinese idiom cloze test & 80 \\
chid\_baidu~\cite{chid} & Baidu Chinese idiom dataset & 80 \\
chinese\_safety\_test\_bias~\cite{chinesesafety} & Safety and bias evaluation & 80 \\
cmrc~\cite{cmrc} & Chinese machine reading comprehension & 80 \\
cola~\cite{cola} & Linguistic acceptability corpus & 77 \\
commonsense\_qa~\cite{commonsenseqa} & Commonsense question answering & 80 \\
conll2014~\cite{conll2014} & Grammatical error correction & 80 \\
counting\_and\_probability~\cite{mmlu} & Math combinatorics problems & 80 \\
GAOKAO-2023\_Math\_en~\cite{gaokao} & Chinese college entrance exam math & 80 \\
gsm8k\_test\_100~\cite{gsm8k} & Grade school math (subset) & 80 \\
IFEval~\cite{ifeval} & Instruction following evaluation & 320 \\
lcsts~\cite{lcsts} & Chinese short text summarization & 79 \\
LiveCodeBench100-2305-2409~\cite{livecodebench} & Live programming evaluation & 80 \\
math~\cite{math} & General math problems & 80 \\
math23k\_test\_100~\cite{math23k} & Math word problems (subset) & 80 \\
MATH-500~\cite{math} & Challenging math competition problems & 400 \\
mmlu\_pro~\cite{mmlupro} & Massive multi-task understanding & 224 \\
msra~\cite{msra} & Named entity recognition & 80 \\
newstest2020~\cite{newstest2020} & Machine translation evaluation set & 80 \\
nlpcc2017\_task2~\cite{nlpcc2017} & News headline categorization & 80 \\
nlpcc2018\_task2~\cite{nlpcc2018} & Grammatical error correction & 80 \\
number\_theory~\cite{mmlu} & Math number theory problems & 80 \\
ocnli~\cite{ocnli} & Chinese natural language inference & 80 \\
prealgebra~\cite{mmlu} & Pre-algebra math problems & 80 \\
precalculus~\cite{mmlu} & Precalculus math problems & 80 \\
qnli~\cite{squad} & Question-answering NLI & 80 \\
qqp~\cite{qqp} & Quora question pairs similarity & 80 \\
race~\cite{race} & Reading comprehension & 80 \\
reco~\cite{reco} & Chinese reading comprehension & 80 \\
squad~\cite{squad} & Reading comprehension & 80 \\
sst2~\cite{sst2} & Sentiment analysis (binary) & 80 \\
story\_cloze\_test~\cite{storycloze} & Story completion and reasoning & 80 \\
sts\_b~\cite{stsb} & Semantic textual similarity benchmark & 80 \\
truthful\_qa~\cite{truthfulqa} & Truthfulness evaluation in QA & 80 \\
webqa~\cite{webqa} & Web-based question answering & 80 \\
winogrande\_wsc~\cite{winogrande} & Coreference resolution & 80 \\
\midrule
\multicolumn{2}{c|}{Total} & 4596 \\
\bottomrule
\end{tabular}
\caption{The query pool used to train the scorer for the larger agent pool.}
\label{tab:large_data_train}
\end{table*}

%% file: tex_files/layer1_prompt.tex
\begin{figure}[htbp]
\begin{promptbox}
You are participating in a multi-agent reasoning task.\\

**Your objectives**\\
1. Produce the best possible answer to the user's query.\\
2. Critically evaluate your own answer and give it a quality score **between 0 and 1**  \\
    (0 = completely wrong, 1 = perfect).\\

**Output format** - return **ONLY** a valid JSON object:\\
```json\\
\{\\
    "answer": "<your answer>",\\
    "self\_score": <float between 0 and 1>\\
\}\\
'''\\

Do **not** add any keys, comments or extra text.
\end{promptbox}
\caption{The prompt used in layer-1.}
\label{prompt:layer1}
\end{figure}

%% file: tex_files/layer2_prompt.tex
\begin{figure}[htbp]
\begin{promptbox}
You are participating in a multi-agent reasoning task.Here are several answers from other LLMs:\\

{answer\_block}\\

**Your objectives**\\
1. Taking every answer in the previous round into account and produce an improved answer to the user's query.\\
2. Critically evaluate your own answer and give it a quality score **between 0 and 1**  \\
    (0 = completely wrong, 1 = perfect).\\
3. Critically evaluate **each** ANSWER\_i above with a value in [0, 1] representing its quality.\\
    (0 = completely wrong, 1 = perfect)\\

**Output format** - return **ONLY** a valid JSON object:\\
```json\\
\{\\
    "answer": "<your improved answer>",\\
    "self\_score": <float>,\\
    "peer\_scores": [<float\_score\_for\_ANSWER\_0>, <float\_score\_for\_ANSWER\_1>, ...]\\
\}\\
'''\\
Do not include any other text.
\end{promptbox}
\caption{The prompt used in intermediate layer-$i|_{1<i<l}$.}
\label{prompt:layer2}
\end{figure}

%% file: tex_files/layer3_prompt.tex
\begin{figure}[htbp]
\begin{promptbox}
You have been provided with a set of responses from various open-source
models to the latest user query. Your task is to synthesise these
responses into a single, high-quality answer. Critically evaluate the
information given, correct any mistakes, and produce a coherent,
well-structured response that meets the highest standards of accuracy.\\

Responses from models:\\
1.{model\_response\_1}\\
2.{model\_response\_2}\\
...
\end{promptbox}
\caption{The prompt used in the last layer.}
\label{prompt:layer3}
\end{figure}

%% file: tex_files/data_gen_prompt.tex
\begin{figure}[htbp]
\begin{promptbox}
You have been provided with a set of responses from various open-source models to the latest user query. Your task is to synthesize these responses into a single, high-quality response. It is crucial to critically evaluate the information provided in these responses, recognizing that some of it may be biased or incorrect. Your response should not simply replicate the given answers but should offer a refined, accurate, and comprehensive reply to the instruction. Ensure your response is well-structured, coherent, and adheres to the highest standards of accuracy and reliability.\\

This is the original question answered by these models:\\
{original question}\\

Responses from models:\\
1.{model\_response\_1}\\
2.{model\_response\_2}\\
...
\end{promptbox}
\caption{The prompt used to generate model response with reference answers.}
\label{prompt:data_gen}
\end{figure}

%% file: tex_files/single_prompt.tex
\begin{table*}[htbp]
\centering
\caption{The prompt of each dataset.}
\label{prompt:single}
\resizebox{\textwidth}{!}{
\begin{tabular}{l|p{14cm}}
\toprule
Dataset & Prompt \\
\midrule
 MATH & Answer the following multiple choice question. The last line of your response should be of the following format: 'ANSWER: \$LETTER' (without quotes) where LETTER is one of ABCD. Think step by step before answering.
\{question\}
\{options\} \\
\midrule
 GSM8k & \{question\}
Please reason step by step, and put your final answer within \textbackslash boxed\{\}.\\
\midrule
 ARC-c & \{problem\}
Please reason step by step, and put your final answer within \textbackslash boxed\{\}.\\
\midrule
 MBPP & You are an expert Python programmer, and here is your task:
\{prompt\}
 Your code should pass these tests:
\{test\_list\}\\
\midrule
 RACE-high & Answer the following multiple choice question. The last line of your response should be of the following format: 'ANSWER: \$LETTER' (without quotes) where LETTER is one of ABCD. Think step by step before answering.
\{question\}\{options\}\\
\midrule
 MMLU-biomed & Answer the following multiple choice question. The last line of your response should be of the following format: 'ANSWER: \$LETTER' (without quotes) where LETTER is one of ABCD. Think step by step before answering.
Article: \{article\}
Q:\{questions\}
\{options\}\\
\bottomrule
\end{tabular}
}
\vspace{-10pt}
\end{table*}